\documentclass{article}
\pdfoutput=1
\PassOptionsToPackage{numbers, compress}{natbib}
\usepackage{enumitem}

\usepackage[final]{nips_2016}

\usepackage[utf8]{inputenc} % allow utf-8 input
\usepackage[T1]{fontenc}    % use 8-bit T1 fonts
\usepackage{hyperref}       % hyperlinks
\usepackage{url}            % simple URL typesetting
\usepackage{booktabs}       % professional-quality tables
\usepackage{amsfonts}       % blackboard math symbols
\usepackage{nicefrac}       % compact symbols for 1/2, etc.
\usepackage{microtype}      % microtypography
\usepackage{tablefootnote}

\usepackage{amsmath}
\usepackage{bm}
\usepackage{graphicx} % TODO: This is ok, right?
\setkeys{Gin}{draft=false} % TODO: Why is this needed to show images, even in  final mode?

\title{Convolutional Residual Memory Networks}

\author{
  Joel Moniz \& Christopher Pal \\
  Polytechnique Montréal \& \\
  The Montréal Institute for Learning Algorithms\\
  Université de Montréal \\
  \texttt{christopher.pal@polymtl.ca} \\
}

\begin{document}

\maketitle

\begin{abstract}
Very deep convolutional neural networks (CNNs) yield state of the art results on a wide variety of visual recognition problems. A number of state of the the art methods for image recognition are based on networks with well over 100 layers and the performance vs. depth trend is moving towards networks in excess of 1000 layers. In such extremely deep architectures the vanishing or exploding gradient problem becomes a key issue. Recent evidence also indicates that convolutional networks could benefit from an interface to explicitly constructed memory mechanisms interacting with a CNN feature processing hierarchy. Correspondingly, we propose and evaluate a memory mechanism enhanced convolutional neural network architecture based on augmenting convolutional residual networks with a long short term memory mechanism. We refer to this as a convolutional residual memory network. To the best of our knowledge this approach can yield state of the art performance on the CIFAR-100 benchmark and compares well with other state of the art techniques on the CIFAR-10 and SVHN benchmarks. This is achieved using networks with more breadth, much less depth and much less overall computation relative to comparable deep ResNets without the memory mechanism. Our experiments and analysis explore the importance of the memory mechanism, network depth, breadth, and predictive performance. 
%
%Furthermore, we present a siamese variant of this architecture for comparing images along with results for face verification. Here again we find that convolutional neural networks augmented with this memory mechanism yield increased performance. 
\end{abstract}

\section{Introduction}

Deep convolutional neural networks have recently been found to exceed human level performance on a number of tasks. As an example, residual networks (ResNets) have taken the top spot on the standard ImageNet evaluation, exceeding human performance in terms of the key top-5 error rate \cite{he2015deep}.
% CP note double check if this is the right ref.
%
Recently, considerable progress has been made towards making deep architectures significantly deeper \cite{simonyan2014very, szegedy2015going, srivastava2015training, he2015deep, he2016identity}. % Should I cite their newest paper?
These extremely deep networks have demonstrated their ability to learn higher level features, which in turn generalize more effectively for classification tasks. 
% Further, a deeper network works much better when being trained by extremely large dataset.

Classical designs for deep convolutional neural network architectures such as the well known AlexNet \cite{krizhevsky2012imagenet} and VGG networks \cite{simonyan2014very} do not have explicitly designed memory mechanisms. However, as visual recognition networks become deeper, the ability to interface with a memory mechanism and to manipulate memory contents outside of the traditional CNN processing pipeline becomes a seductive property.  

In contrast, explicitly constructed memory mechanisms for recurrent neural networks (RNNs) have been known and actively explored for many years. In particular, the Long Short Term Memory (LSTM) mechanism \cite{hochreiter1997long} has emerged as a seminal contribution, demonstrating state of the art performance on a wide variety of problems. They also offer a solution to the vanishing or exploding gradient problem in RNNs with many time steps. However, RNN and LSTM approaches have traditionally been applied to and have excelled at sequential data such handwriting recognition, speech recognition, text analysis and generation and video. 

The residual network approach of \cite{he2015deep} uses a simple skip connection mechanism to propagate information to higher levels of a network more readily. In contrast, recently proposed Highway Networks \cite{srivastava2015training} have a more sophisticated interface between a convolutional network and a memory mechanism. We discuss these methods in further detail in our Related Work section below.

In our work here we explore a novel memory enhanced deep convolutional network architecture. In our approach, 
instead of using a deep LSTM network as a mechanism that processes elements in a time series or in a specific pattern or sequence, its input are the abstract feature representations of a very deep CNN. In this way the LSTM mediates a memory interface between the internal memory and the algorithmic state of an LSTM RNN with the features created by the layers of a CNN.  In this scheme each input step represents a higher level of abstraction, but lower level abstractions can be remembered and manipulated as needed by the LSTM RNN. %Thus, in our proposed architecture an LSTM acts as a component that can be directly interfaced to a state of the art deep CNN architecture. 
We focus on exploring this type of memory interface with an LSTM memory manipulation and algorithmic execution architecture using deep residual networks as the underlying base. We refer to this specific type of network as convolutional residual memory networks CRMN and a general CNN architecture enhanced with memory manipulation capabilities as a convolutional memory network. However, interestingly, as our formulation is based on an LSTM, the resulting architecture also provides a parallel path of network computation for CNNs extended in this way. This affords such models the capacity to perform both memory manipulation as well as providing alternative general purpose algorithmic processing operations that execute along side of classical convolutional network processing units.

We show an image of our architecture in Figure \ref{fig:monorail}. The LSTM executes and interfaces along side the underlying residual CNN architecture, processing the successively more abstract, higher level feature representations output from the convolutional processing units of the CNN. The output from both the LSTM and the original network are then used as the input feature representation given to the final classification layer(s) of the network. We describe this model in more detail in Section 3 below. 

% Talk about varying output sizes not being an issue?

%This section assumes that the reader is familiar with Long-Short Term Memory Networks. The author personally finds \cite{colahlstm} an excellent resource that explains the concept of LSTMs extremely clearly and intuitively.

%Image classification has a wide range of practical applications, such as automatic categorization of an object, and face recognition. Earlier, classification often involved carefully hand-picked, fine-tuned features, often determined with the help of domain experts. But with the advent of deep learning, significant progress has been made towards have the system automatically determine the ideal features to use with almost no user intervention.

\section{Related Work}

\subsection{Training Very Deep Convolutional Networks} %Add in diagrams and our model being a generalization of this model?

As deep networks optimized with gradient descent must deal with the issue of vanishing or exploding gradients, various architectures and mechanisms have been proposed to address the underlying issue. For example, both deep residual networks and deep networks using exponential linear units or ELUs have recently emerged as among the top performing architectures on the CIFAR-10 and 100 benchmarks that we examine here. These two top performing techniques on the standard CIFAR benchmarks use different methods to deal with the underlying gradient problem. 

In the first case, residual connections composed of skip connections that are combined additively with the output of convolutional blocks of are used. In the second case exponential linear units are used, which, like rectified linear units have the property that gradients of the activation function with respect to their input are equal to one for inputs in excess of zero. 

Recurrent neural networks (RNNs) also suffer from similar issues with vanishing and exploding gradients. LSTM RNNs were explicitly designed to deal with the issue through the manipulation of a memory cell having the property that gradient information can be captured and stored without degradation over many timesteps -- or in our case, many network layers. 

Recurrent convolutional networks (RCNs) have been used for video processing \cite{srivastava2015unsupervised,donahue2014long}. These approaches typically feed information from 2D CNNs into an RNN which creates the representation for the video. 
However, a recently proposed method for video processing \cite{ballas2015delving} has integrated CNNs with gated recurrent units or GRUs \cite{chung2014empirical} both temporally and along the feature map hierarchy to form a stacked GRU-RCN. They use the output of a VGG CNN architecture that has been pretrained on ImageNet \cite{simonyan2014very} then finetuned on an activity recognition dataset. They then use the top four pooling layers and one fully connected layer as input to a stacked GRU-RCNs. The network is therefore potentially very deep in terms of the temporal dimension, but at 5 layers it is shallow in terms of per frame abstraction depth. The positive impact of having a GRU recurrent network along the abstraction depth of a VGG grade CNN points to potential advantages of using more sophisticated RNN mechanisms along the feature abstraction hierarchy of extremely deep CNN architectures.

%Very simply, 
The recently proposed Highway Network \cite{srivastava2015training} architecture consists of a mechanism allowing 2D-CNNs to interact with a simple memory mechanism. The details of the memory mechanism are of course very important. In a Highway Network input layers $\textbf{x}_l$ are transformed to $\textbf{h}_l$ using a typical neural network affine transformation followed by an activation function. In the case of a convolutional highway network this is implemented as a layer of convolutional units. Elements of the convolutional layer $\textbf{x}_l$ are either passed or blocked and replaced by the transformed $\textbf{h}_l$ using a gating mechanism controlled by a \emph{Transform Gate}, $\textbf{t}^g_l$ and a \emph{Carry Gate}, $\textbf{c}^g_l$ respectively. The final output is the a sum of the content modulated by these gates, 
\begin{align*}
\textbf{x}_{l+1}&=\textbf{h}_l(\textbf{x}_l,\textbf{W}_h)\odot\textbf{t}^g_l(\textbf{x}_l,\textbf{W}_l)+\textbf{x}_l \odot \textbf{c}^g_l(\textbf{x}_l,\textbf{W}_c).
\end{align*}
The gates themselves are learned functions of the input where in the simple highway network this could be a sigmoid neuron with weights given by $\textbf{W}_l$ and $\textbf{W}_c$. However for simplicity \cite{srivastava2015training} uses $\textbf{c}^g_l=\textbf{1}-\textbf{t}^g_l$. Convolutional highway layers use convolutional layers (i.e. weight sharing and local receptive fields) to control the transform gate and to perform the transform itself, using $\textbf{W}_h$. Based on the values of the gates, the output of each block can be the output of the convolutional layer, the same as the input, or a mix of both. Initializing the biases of the transform gate to a negative value initially effectively makes several layers disappear, in a manner of speaking, since their values are not passed though. Highway Networks were found to not suffer from increasing depth, and converge much better than a regular deep CNN. This work further substantiates the notion that explicit memory mechanisms interacting with CNNs could be a potent combination.

\begin{figure}[htbp]
  \centering
    \includegraphics[scale=0.25]{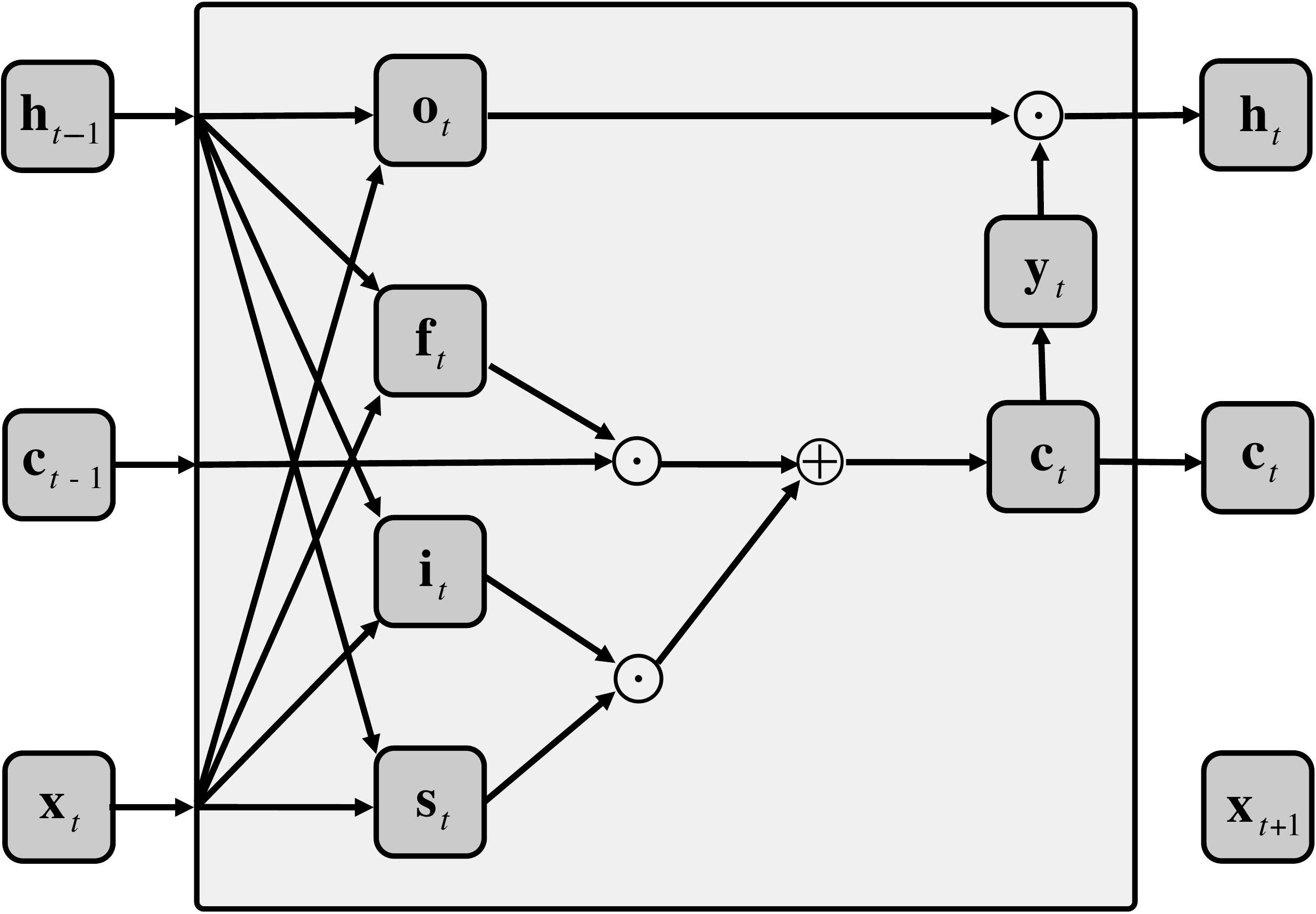} \hspace{.2cm}
     \includegraphics[scale=0.25]{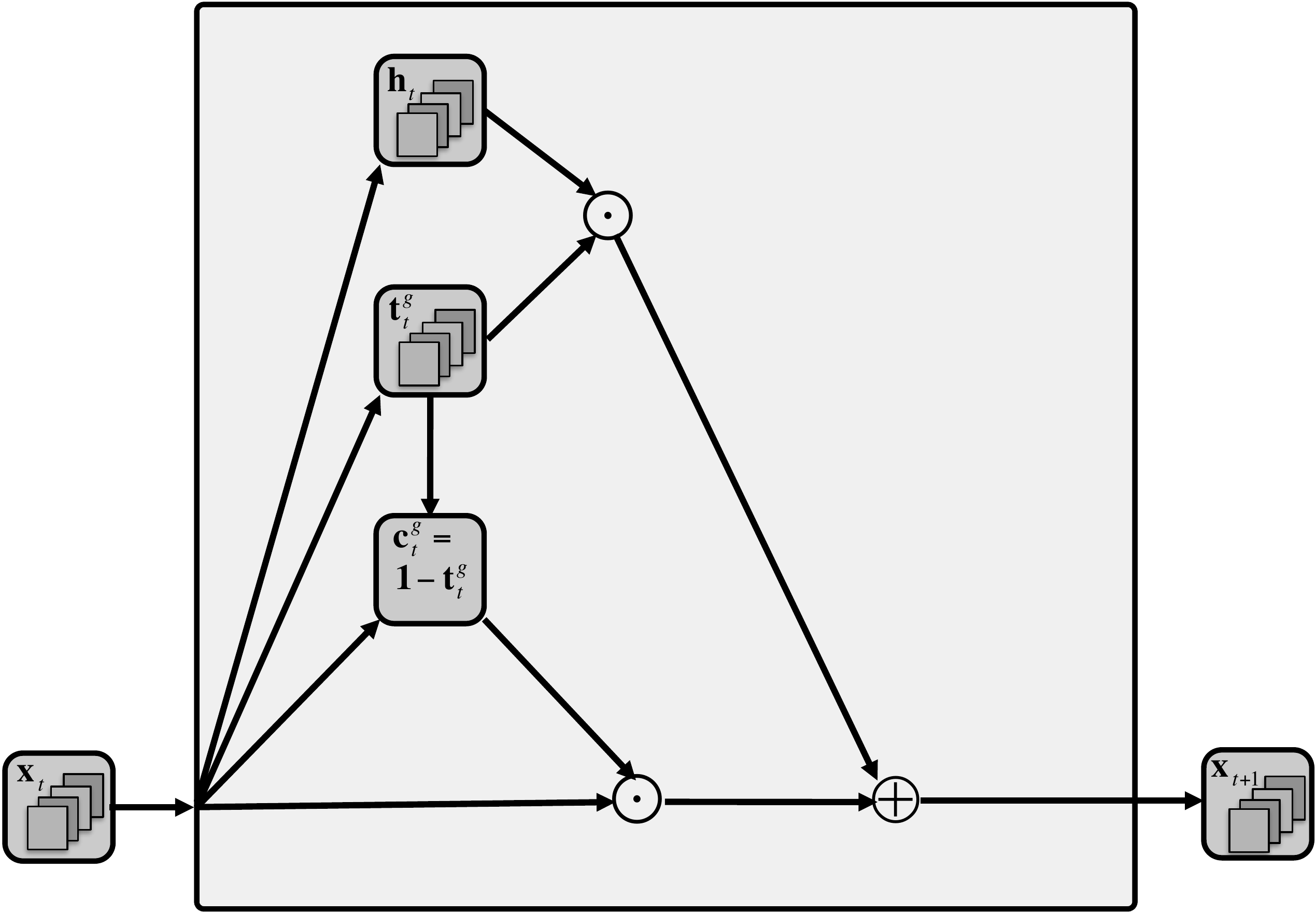}
  \caption[The CRMN architecture]{(Left) An LSTM computing block and (Right) a convolutional Highway network computing block.}
  \label{fig:lstmHighway}
\end{figure}

It is useful to compare convolutional highway networks discussed above with standard LSTMs. We show a visual comparison in Figure \ref{fig:lstmHighway}. We follow the standard conventions for defining LSTMs, see \cite{graves2005framewise} for the explanation of each element in detail. Note we do not show peephole connections; however we use them for our experiments. We discuss the way in which we use an LSTM to interact with a ResNet in more detail in Section \ref{sec:arch} below and the computational processing elements in Figure \ref{fig:lstmHighway} can be contrasted with our convolutional residual memory network approach shown in Figure \ref{fig:monorail}.

\subsection{Residual Networks}

Residual Networks\cite{he2015deep} are based on the use of shortcut connections that skip blocks of convolutional processing layers which are simply recombined with the output of convolutional blocks using addition. The method was motivated by the observation that very deep networks failed to train- the training error \emph{increased} after a certain point. However, in the worst case, there ought to exist at the very least an identity mapping which would guarantee that the deep network does not perform \emph{worse} than a shallow network. The residual networks of \cite{he2015deep} have VGG Net style blocks of 2 convolutional layers each with \(3\times3\) filters. The output of each block is then added with that of the previous block (which is the identity mapping/short-cut connection). The argument is that in this way, it is easier to learn small displacements from 0 than to first learn the identity mapping and then learn a displacement from that mapping. In this way, they successfully trained a network over a hundred layers deep. Their architecture also avoids explicit pooling layers altogether, achieving sub-sampling by simply using convolution layers of stride 2.

%\subsection{Identity Mappings in Deep Residual Networks}

In the pre-activation variant of ResNets \cite{he2016identity}, they try various architectural variant of the earlier proposed ResNets, such as skip connections and gating, and show what effects and advantages using identity mapping has. They show that for easier optimization, an identity after-addition path is advantageous, and this can be achieved by using the activation function \emph{before} the convolutional layer. They also find that using batch normalization, but before the activation so as to maintain the identity, causes the network to train much more easily than, and to outperform the performance of the earlier formulation for a ResNet.
% Mention Wide ResNets here?

\section{Convolutional Residual Memory Networks}
\subsection{Overview}
At a conceptual level, classical implementations of deep convnets potentially have difficulty seeing "the big picture". Although more recent architectures like Highway Networks and ResNets make significant progress with respect to helping information flow along the network, they do this primarily from the ease of optimization stand-point. The CRMN architecture we propose below essentially drops an LSTM on top of a standard convnet architecture (in this case, a ResNet). The LSTM then takes views of the abstraction hierarchy as input, instead of a time series, with features increasing in abstraction as the series progresses (as opposed to changing in time). While variants of these high level notions have been explored in the past, to our knowledge they have not been explored with both extremely deep architectures and the sophisticated memory manipulation and algorithmic execution properties of LSTMs. 

\subsection{Architecture}
\label{sec:arch}
\begin{figure}[htbp]
  \centering
    \includegraphics[width=\linewidth,height=\textheight,keepaspectratio]{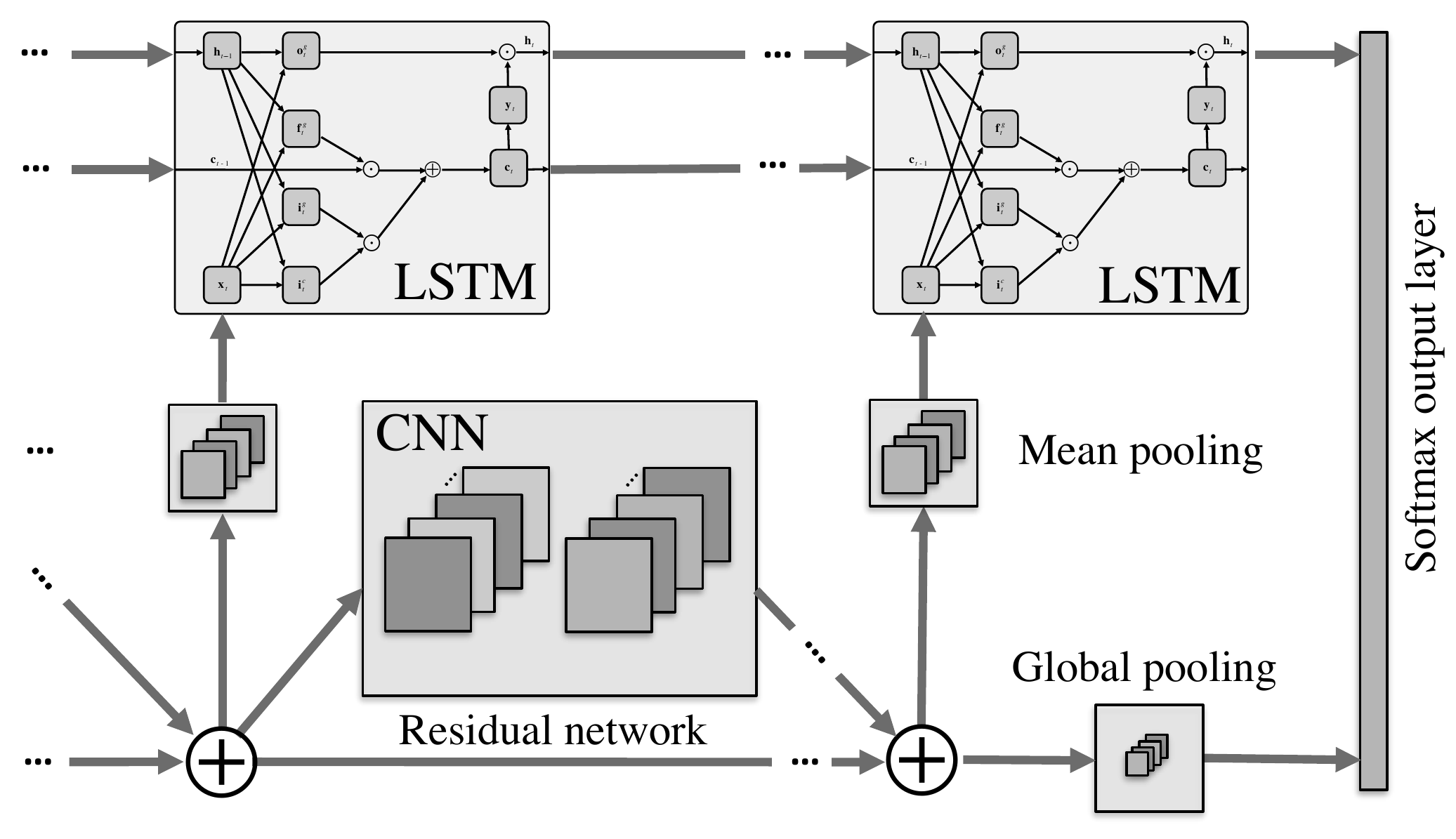}
  \caption[The CRMN architecture]{The repeating computational block and the final computational units of a convolutional residual memory network (CRMN).}
  \label{fig:monorail}
\end{figure}

Our formulation of a CRM Network interfaces a residual network and an LSTM as shown in Figure \ref{fig:monorail} and can be understood as follows. The features at each output block of the residual network just after the non-linearity is applied and the residual shortcut has been added, are fed as the input to the LSTM. The first convolutional layer and the last global pooling 
layer of the ResNet are, however, not used. Thus, for a ResNet with \(6\times{n}+2\) layers, the LSTM is fed a total of \(3\times{n}\) blocks of features. %feature steps, data steps, what do i call these?
However, to cut down the number of parameters and significantly improve performance, before passing each feature map into the LSTM, we apply a \(2\times{2}\) meanpool sub-sampling on the feature map. The final hidden state output of the LSTM, along with the output of the global average pooling layer at the end of the ResNet %clear and techincally sound?
is then fed into a softmax layer. In this architecture, the LSTM merely feeds off of the data output by each residual block, and does not feed anything back into the ResNet pipeline. The size of the total number of features after each block changes twice in a ResNet (each time a convolution of stride 2 is used). When the total number of features is less than the maximum number, the features are zero-padded.
%Zero padding doesn't sound all that convincing

% TODO Is the figure inserted correctly?

%\section{Equations}
%\subsection{Convolutional Residual Memory Networks}

The explanation above can be made more precise as follows. We begin with a ResNet computational block following the formulation in \cite{he2015deep}. Define $\boldsymbol{x}_l$ as the set of all convolutional feature maps for layer $l$ and $\boldsymbol{W}_l$ as all parameters for the layer. We use the output of a ResNet with 2-layer convolutional blocks denoted by the residual mapping function \(\boldsymbol{f}(\boldsymbol{x}_l, \{\boldsymbol{W}_l\})\)) along with the identity mapping as the input to each LSTM unit after the element-wise addition with the previous layer. This implies the following formulation for our CRMN, where
%
%\begin{align} 
\begin{equation}
\boldsymbol{x}_l = \boldsymbol{x}_{l-1}+\boldsymbol{f}(\boldsymbol{x}_{l-1}, \{\boldsymbol{W}_l\}), \\
\end{equation} 
is the input at each LSTM interface, step $l$. As discussed above, we reduce the size of this input using mean pooling. Using  $\boldsymbol{W}_x$ and $\boldsymbol{b}_x$ to denote the various weight matrices and bias terms, the input, forget and output gates for the LSTM are then given by
\begin{align}
\boldsymbol{i}_l &= \boldsymbol{\sigma}_{iof}(\boldsymbol{W}_{xi} \boldsymbol{x}_l  + \boldsymbol{W}_{hi} \boldsymbol{h}_{l-1} + \boldsymbol{w}_{ci} \odot \boldsymbol{c}_{t-1} + \boldsymbol{b}_i), \\
% \end{equation}
% \begin{equation}
\boldsymbol{f}_l &= \boldsymbol{\sigma}_{iof}(\boldsymbol{W}_{xf} \boldsymbol{x}_l + \boldsymbol{W}_{hf} \boldsymbol{h}_{l-1}
       + \boldsymbol{w}_{cf}  \odot \boldsymbol{c}_{l-1} + \boldsymbol{b}_f), \\
       \boldsymbol{o}_l &= \boldsymbol{\sigma}_{iof}(\boldsymbol{W}_{xo} \boldsymbol{x}_l + \boldsymbol{W}_{ho} \boldsymbol{h}_{l-1} + \boldsymbol{w}_{co} \odot \boldsymbol{c}_l + \boldsymbol{b}_o), 
% \end{equation}
% \begin{equation}
\end{align}
where we use the \(logistic\) function as the nonlinearity for the input, forget and output gates \(\boldsymbol{\sigma_{iof}}\). %
The peephole connections are given by the terms of the form $\boldsymbol{w}_{x} \odot \boldsymbol{c}_t$. 
The cell gates are given by
\begin{align} 
\boldsymbol{c}_l &= \boldsymbol{f}_l  \odot \boldsymbol{c}_{l - 1}
       + \boldsymbol{i}_l \odot
       \boldsymbol{s}_l, \\
\boldsymbol{s}_l &=        \boldsymbol{\sigma}_c(\boldsymbol{W}_{xc} \boldsymbol{x}_l + \boldsymbol{W}_{hc} \boldsymbol{h}_{l-1} + \boldsymbol{b}_c), 
\end{align}
where we use \(tanh\) as the cell gate non-linearity \(\boldsymbol{\sigma_{c}}\). The final output for the hidden unit update is given by
\begin{equation}
\boldsymbol{h}_l = \boldsymbol{o}_l \odot \boldsymbol{\sigma}_h(\boldsymbol{c}_l).
\end{equation}
As discussed above and illustrated in figure \ref{fig:monorail}, the final prediction is made using global pooling of the final ResNet layers and the output of the LSTM hidden state as input to a softmax prediction layer.
%
%While $\boldsymbol{X}$ represents a feature output from a residual block, $\boldsymbol{x}$ represents the input to the LSTM, $\boldsymbol{X}$ flattened into a vector.
%
%\subsection{A comparison to highway networks}
%
%It is useful to compare Convolutional Highway networks with our formulation above. Given a canonical neural network activation function producing output $$ In a highway network on replaces the usual neural network activation function with a transform gate T and a carry gate C
%
%
\subsection{Complexity Considerations}

Although the memory manipulation mechanism of the LSTM applied on top of the residual network adds parameters and computational complexity the impact is moderate.
In terms of the number of flops, the fact that our formulation does not use convolutions within the gating control and input mechanism implies a lower computational load at the memory interface to the LSTM. Consequently, in practice a 32 layer memory enhanced ResNet model requires far fewer FLOPs compared to a base ResNet with a similar number of parameters.
%
%The overall computational complexity remains the same with the memory network as without it. 
%

%
Consider, for the sake of simplicity, each convolutional block as comprising of a 2 convolutional layers of size \(n\times{n}\), with filter size \(k\times{k}\), \(f\) feature maps and a stride of 1, with an element-wise residual addition at the end. The total number of operations (with multiplications, additions and the activation function each being considered as an operation) done by each block is then given by:
\begin{align}
% \begin{equation*}
\begin{split}
    C_r %= 2[(2fk^2-1)\times{fn^2} + fn^2 + fn^2] + fn^2 
%    & = (4fk^2 + 3)fn^2 \\
     = \mathcal{O}(f^2k^2n^2).
\end{split}
% \end{equation*}
%
\intertext{In the proposed architecture, the output from each block (of size \(i=fn^2\)) is fed as a datastep into the LSTM with \(h\) hidden states, after the residual addition. Thus, the cost becomes}
%
% \begin{equation*}\label{eq:crm1}
% \label{eq:crm1}
\begin{split}
    C_{rm} %& = C_r +  3[2(2h-1)h + (2i-1)h + h + h] + [{(2h-1)h + (2i-1)h + h + h + h} + h] + 2h \\
%           & = C_r +  h[14h+8i+1] \\
            = \mathcal{O}(f^2k^2n^2 + h^2 + hi).
\end{split}
% \end{equation*}
%
\intertext{If, as in our experiments, the number of hidden states is far fewer than the number of features in any given convolutional layer, i.e., if \(h<<i\), we have}
%
% \begin{equation*}
\begin{split}
    C_{rm} = \mathcal{O}(f^2k^2n^2 + hfn^2) 
            = \mathcal{O}(fn^2(fk^2 + h)).
\end{split}
% \end{equation*}
\end{align}
Furthermore, it is an important aspect of our formulation that the additional parameters introduced by adding the memory network are constant, independent of the depth of the architecture. 

\section{Experiments}

We have experimented with many variants of the proposed architecture. Some variants cut down the number of parameters -- including using a GRU \cite{cho2014properties} instead of an LSTM for example, or using the bottleneck variants of ResNets and pre-activation ResNets. We also examined variants where the residual connections were not used. However, all these variants were found to not perform as well on the validation set used. We therefore present a subset of our most informative experiments below.

%\subsection{Pre-processing and Training}

%\subsection{Training} \label{monorailtrain}
In order to avoid over-fitting the model to the test set, the architecture and its hyper-parameters were chosen based on a 10\% subset of the training set, used as a validation set. %Training is performed as follows:
%\begin{itemize}[noitemsep,topsep=2pt]
%\setlength\itemsep{.5em}
    %\item 
    An initial learning rate is chosen (\(\alpha\)=0.1), along with a learning rate schedule, representing how the initial learning rate would be changed.
    %\item 
    The model is trained until neither the validation error nor the validation accuracy improves for \(\beta\) epochs, at which point the previous best model is loaded and the system trained with a lower learning rate. This is repeated until the learning rate falls below a set threshold.
    %\item 
    The learning schedule so obtained is then used on another run with all the training data included, and the test error of the model trained with this schedule reported.
%\end{itemize}
%
% Note that although reaching the "optimum" training error using the lowest learning rate in the schedule when training with the validation data may take longer, it enables reaching the error value without over-shooting it, which would defeat the purpose of early stopping.

% Once the learning schedule is determine as above, the model is then retrained, changing the learning rate according to the schedule at the optimum epoch as determined above. %not exactly at the optimum epoch, but a little after- how to quantify this?

%\subsection{Initialization}
All layers in the ResNet are initialized exactly as proposed in \cite{he2015deep} (using the initialization scheme proposed in \cite{he2015delving}), using rectified linear units as the activation function. The weights of all the gates in the LSTM are initialized using the scheme proposed in \cite{saxe2013exact}. The input and forget gates of the LSTM use the \emph{sigmoid} non-linearity, while the output gate uses \emph{tanh}. As has already been discovered in \cite{srivastava2015training}, high negative biases tend to be favourable. In this case, this is most likely because a high negative bias initially is going to bias the LSTM to assign more importance to things seen more recently, which is particularly useful because more information is likely to be present at the higher layers, since they represent more abstract features. The biases of each convolutional layer is initialized to 0. The initial hidden state of the LSTM is learned.

%The number of feature maps used seems to have a significant impact on the classification accuracy for CIFAR-100\cite{krizhevsky2009learning}, although we have not explored this in as much detail as it merits, and have limited ourselves to starting off with 16, 24, 32 and 64 feature maps in the ResNet architecture. 
We have kept both the dimensionality of the cell output as well as the size of the cell memory at a constant value of 100 in the following experiments, and used a momentum of 0.9 and a weight decay of 0.0001 throughout. The entire implementation was done in Theano \cite{2016arXiv160502688short} and Lasagne.
% Wide ResNets here too, instead of citing krizhevsky2009learning?

\subsection{Comparing CRMNs with ResNets}

\begin{table}[htb!]
\centering
\begin{tabular}{ p{1.5cm}  p{.9cm} p{.7cm}  p{1.7cm}  p{1.5cm}  p{1.9cm}  }
 \hline
Model & Layers & F.map 16 $\times$ & Parameters \newline (million) &  Sec. per \newline epoch  & Test set \newline accuracy (\%)   \\ \hline
ResNet	&  134 &	1	& 2.12  &	455  &	71.54 \\ 
ResNet	&  104 &	1.5	& 3.67  &	473  &	71.14 \\ 
ResNet	&  92  &	2	& 5.74  &	583  &	74.39 \\ 
ResNet	&  62  &	4	& 15.16 &	881  &	71.78 \\ \hline 
ResNet  &  32  &  1	    & 0.47 &	125  & 69.53 \\ 
ResNet  &  32  &  1.5   & 1.05 &	173  & 71.88 \\ 
ResNet  &  32  &  2	    & 1.86 &	199  & 73.28 \\ 
ResNet  &  32  &  4	    & 7.41 &	427  & 75.73 \\ \hline 
CRMN    &  32  &	1	& 2.16  &	140 &   70.84 \\ 
CRMN    &  32  &	1.5	& 3.56  &	165 &	73.18 \\ 
CRMN    &  32  &	2	& 5.19  &	233 &   74.67 \\ 
CRMN    &  32  &	4	& 14.01 &	490 &	76.39 \\ \hline % 77.03
\end{tabular}
\caption{Comparison of ResNets and CRMNs with models of similar complexity.}
\label{explore}
\end{table}

We provide three blocks of experiments in Table \ref{explore}. In the last block
we compare our formulation of convolutional residual memory networks (CRMNs) with ResNets having a similar number of parameters in the first block of four experiments. To compare we vary both the depth and number of feature maps of the ResNet baselines. We keep the number of layers fixed at 32 for our CRMNs, and examine the performance on CIFAR-100 for models having 16, 24, 32 and 64 feature maps at the first convolutional layer.  We see that there is only one experiment where a ResNet lacking the memory mechanism outperforms the model with the the mechanism. This occurs in the deepest model with 134 layers which has the fewest feature maps. We conjecture that the LSTM simply does not have enough input to work with in this case. Otherwise we see consistent improvement with addition of the LSTM.  

\begin{figure}[htbp]
  \centering
    \includegraphics[scale=0.38]{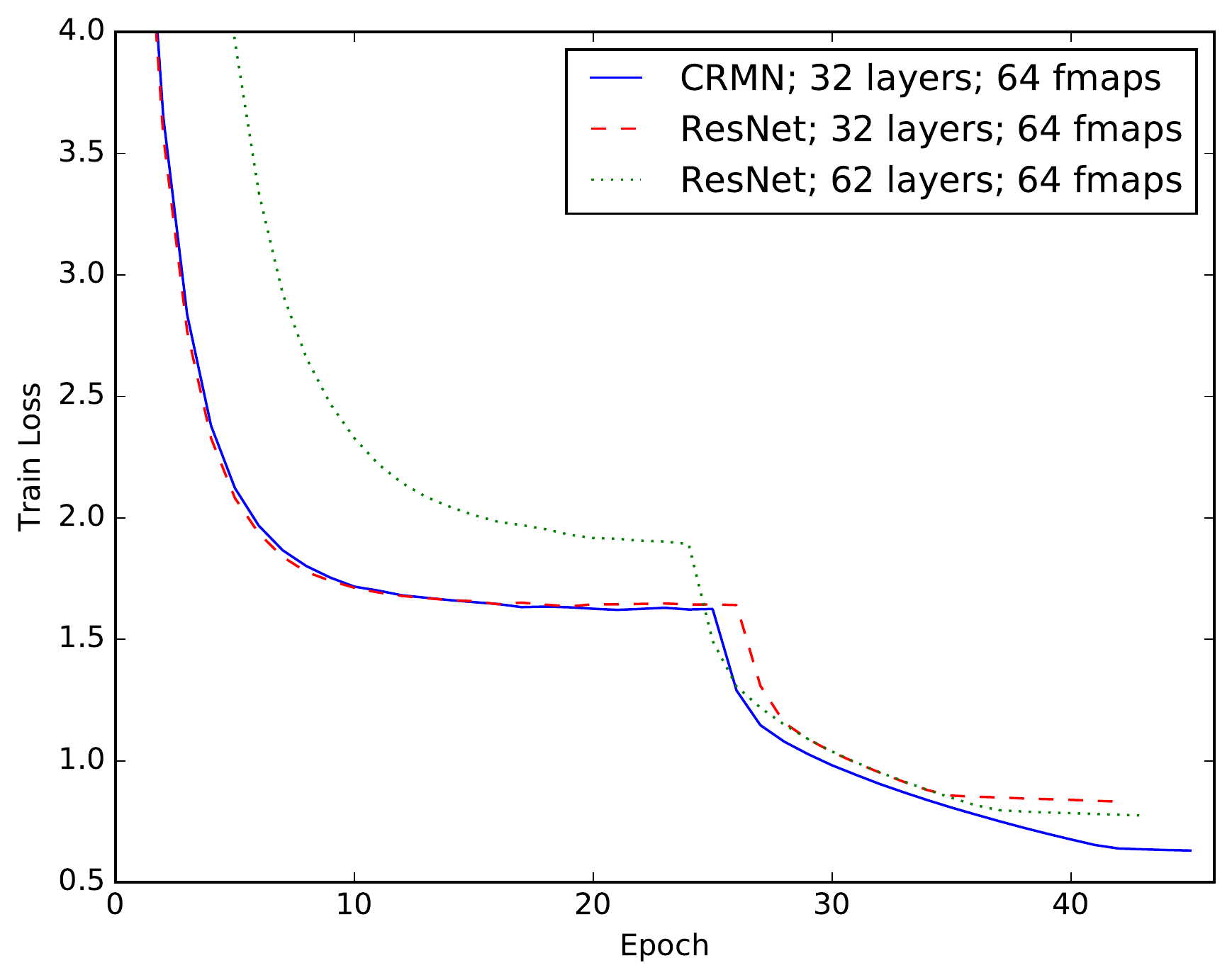} \hspace{.2cm}
     \includegraphics[scale=0.38]{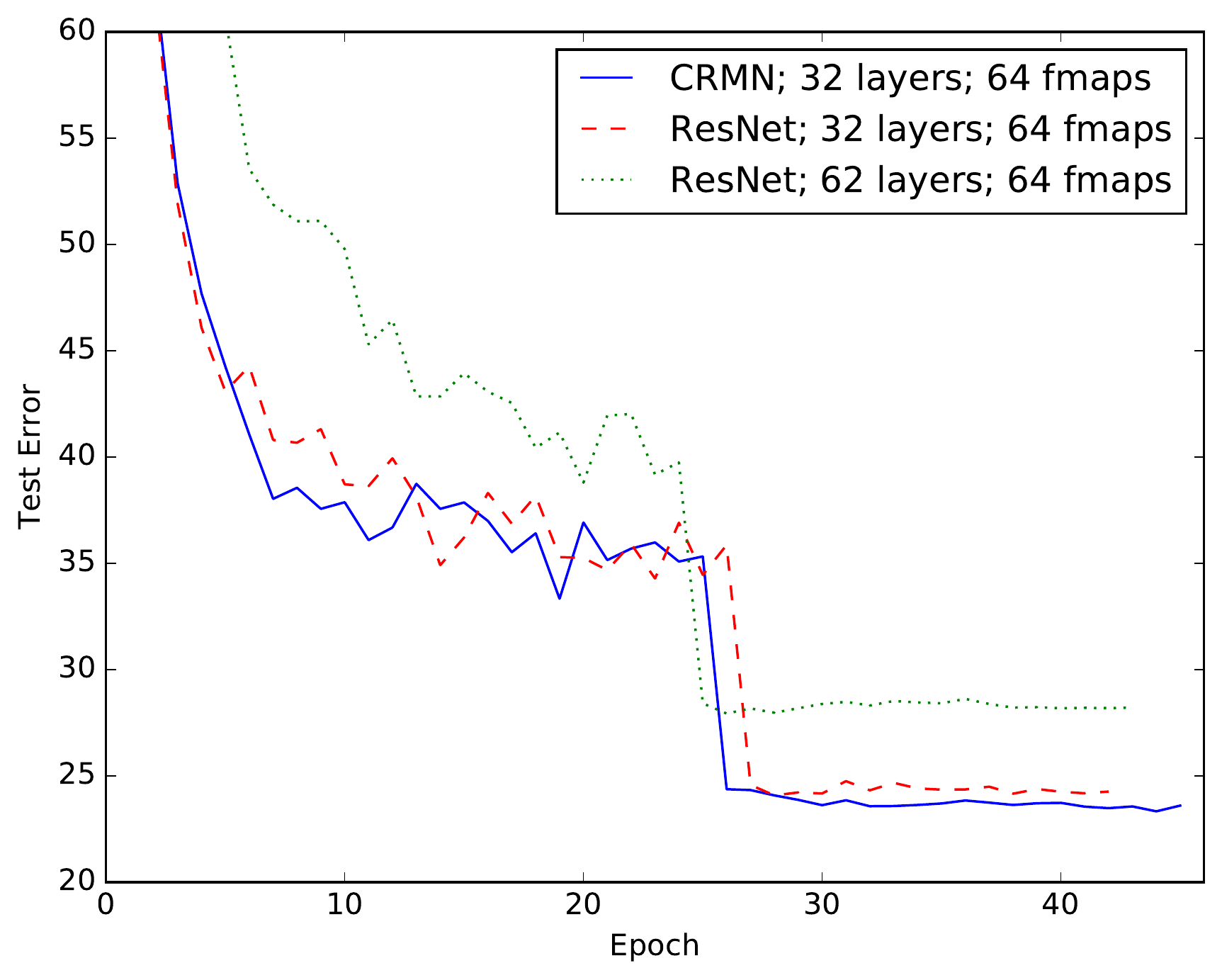}
  \caption[Learning Curves]{(Left) Training set learning curves for similar models (Right) Test set error during final optimization. Models are from the last line of each block in Table \ref{explore}.}
  \label{fig:curves}
\end{figure}

In the second block of Table \ref{explore} we provide a corresponding set of four experiments for ResNets without the memory mechanism, but otherwise having exactly the same architecture as the bottom block of CRMNs. We see uniform improvement over the baselines.
Our CRMN model that had the best validation set performance in Table \ref{explore} corresponded to the final line in the table, or the model with the most parameters. This configuration also yielded the highest test accuracy. 
In Figure \ref{fig:curves} we show training set learning curves and test set error over the final optimization schedule for models in last line of each block in Table \ref{explore}. 
Finally, note also that the LSTM adds a \emph{constant} number of parameters for a given architecture, irrespective of depth. 
%% Also do comparison in terms of FLOPS

\subsection{Comparisons with state of the art for CIFAR-100, CIFAR-10 and SVHN}

We re-ran our model architecture with the highest validation set performance on CIFAR-100 from Table \ref{explore} as well as various extremely wide models %with 28 layers and 160 feature maps 
using a modified batch size (of 50 instead of 100) and learning rate schedule (with learning rates [0.1, 0.01, 0.005, 0.001] instead of [0.1, 0.01, 0.001]). These experiments yielded even higher validation set accuracies and higher test set performance as indicated. Following the reporting protocol in \cite{he2015deep}, over five repetitions of the 32 layer, 64 featuremap configuration our best model yielded 78.27\% accuracy on the test set and the $\textrm{mean} \pm{\textrm{std.}}$ test set performance was 77.65$\pm${0.385}\%. 

We observed that architectures trained for longer before a learning rate shift perform better on the validation set in general, so we train for at least $m$ epochs, before the first learning rate shift, $m$ being a hyperparameter, $m=70$ for all our experiments. Note that we have not optimized this hyperparameter, and it is possible that better results may be obtained if $m$ is made even larger.

The architecture proposed in this paper can be divided into 3 sub-components: the Residual Network (which extracts features that are increasingly abstract as the network becomes deeper), the LSTM (which helps remember less abstract features that are important), and the fully connected softmax layer (which acts as a mediator between the outputs of the ResNet and the LSTM). Correspondingly, we also explored a round robin learning rate (RRLR) schedule when updating the learning rate, wherein we update the learning rates of each component sequentially starting with the ResNet, then the LSTM, then the softmax. Note that although such a system could potentially take much longer to train (since there are more learning rate shifts), we practically observe that far fewer epochs are required between each learning rate update compared to the case when updates to all components are done at the same time.
%(with several components often changing learning rates simultaneously).

We evaluated a model with 28 layers and 160 feature maps using the RRLR schedule. It yielded 79.68\% accuracy on the test set. Another model trained with the RRLR schedule having 32 layers and 192 feature maps yielded 80.21\%, which, to the best of our knowledge, is the first result to cross the 80\% mark for CIFAR-100. The validation set performance for this model was 79.52\% -- the highest we have observed across all our experiments. We evaluated a corresponding ResNet architecture without the LSTM and it yielded 79.30\% on the validation set. %The accuracies listed here are based on runs using early stopping and a learning rate schedule obtained using the validation scheme described above. Further, the network reaches this accuracy after 64 epochs of training.
% Comparing our approach to other prior work on CIFAR-100 reveals that we have obtained results state of the art result while being significantly shallower (i.e. 30 times shallower). The previous best results of which we are a 1001 layer ResNet \cite{he2016identity}\footnote{Appearing on \emph{arXiv} but not peer reviewed and therefore we have not included it in the table.} which yielded 77.29\%. 
A wide Resnet with 28 layers and 160 feature maps was reported to have yielded 79.5\% in \cite{zagoruyko2016wide}, they report that dropout was able to improve performance by another .46\% to 79.96\%. Note that we have not used dropout in our experiments.
% Wide ResNets gives 79.57 with 
% d=28 k=12 params 52.5M 

%% No footnote here? Should $^1$ be removed from the table?
%% Shouldn't we mention that the wider versions use the new ResNet not the old one?
%\footnotetext{Number of feature maps in first convolutional layer}
%Faces: random negative selection; ensured seed was same for comparison

We also note that almost all other methods that we have compared against in Table \ref{comparison} for our CIFAR experiments perform much more pre-processing than we have. While the only pre-processing we perform is to convert each pixel to the range [0,1] and subtract all images by the mean pixel value of the training set (as in \cite{zeiler2013stochastic}), Global Contrast Normalization followed ZCA whitening is fairly standard, as done in \cite{goodfellow2013maxout}.

Experiments on CIFAR-10 using an architecture 28 layers with 160 feature maps and another with 32 layers and 192 feature maps yielded the results shown at the bottom of Table \ref{tab:cifar10}.
% A similar model without the LSTM yielded 94.39\%.
The first configuration was selected to match the layer and feature map structure of \cite{zagoruyko2016wide}. The second configuration of our model was selected through searching over a few architecture configurations and choosing the model with the best validation set performance on CIFAR-100. % Should this read CIFAR-10?

We provide other top performing methods in the table for context. The best results of which we are aware for this benchmark is 96.53\% obtained using the Fractional Max-pooling approach in \cite{graham2014fractional}; however, it was obtained using 100 tests; using a single test the method in \cite{graham2014fractional} yielded 95.5\%. 95.38\% has been reported using a 1001 layer ResNet in \cite{he2016identity}. The Wide ResNet work in \cite{zagoruyko2016wide} reports 95.83\% using a model with 28 layers and 160 feature maps.

Note that in Tables \ref{comparison} and \ref{tab:cifar10}, "fm" represents the number of feature maps present in the first convolutional layer of the ResNet. Further, while we have used the original ResNet architecture \cite{he2015deep} for configurations with fewer than 64 feature maps in the first layer, we have used the pre-activation ResNet architecture \cite{he2016identity} for wider configurations.

\begin{table}[!htb]
    \begin{minipage}{.5\linewidth}
      \centering
     \begin{tabular}{ |p{.5cm} p{5.5cm}|}
    \hline
    Acc. \newline (\%) & Method \\ \hline
     67.61 & Highway Network \cite{srivastava2015training} (100 layers) \\ 
     71.14 & Maxout Network \cite{goodfellow2013maxout}  \\ \hline
     72.34 & LSUV Initialization Method \cite{DBLP:journals/corr/MishkinM15} \\
     72.60 & Bayesian Optimization \cite{snoek2015scalable}  \\ \hline
     74.84 & ResNet \cite{he2015deep} (164 layers) \cite{he2016identity}    \\ 
     % 100-24.58=
     75.42 & Stochastic Depth \cite{huang2016deep} \\ \hline
%     75.72 & ELU Network  \cite{clevert2015fast} (18 layers)  \\  
%   CRMN[ours] & 76.60 \\ \hline
% 74.84 25.16 164 layer - numbers are from newer ResNet paper
% 75.67 24.33 164 preactivation
     77.28 & Swapout \cite{singhHF2016swapout} (32 layers; 64 fm) \\ 
     77.29 & Pre-act. ResNet \cite{he2016identity} (1001 layers) \\ \hline 
     79.50 & Wide ResNet \cite{zagoruyko2016wide} (28 layers; 160 fm) \\
     79.96 & As above with dropout  \\ \hline \hline
          78.27 & Our CRMN (32 layers; 64 fm)  \\
     79.68 & Our CRMN (28 layers; 160 fm), RRLR  \\ 
\textbf{80.21} & Our CRMN (32 layers; 192 fm), RRLR  \\ 
\hline  
    \end{tabular}
\caption{CIFAR-100 Accuracies.}
\label{comparison}
    \end{minipage} 
    \begin{minipage}{.5\linewidth}
%      \caption{}
      \centering
      %
       % \begin{table}[h!]
\centering
    \begin{tabular}{ |p{.5cm} p{5.5cm}|}
    \hline
    Acc. \newline (\%) & Method \\ \hline
%    90.62 & Maxout Network \cite{goodfellow2013maxout} \\ \hline
    92.40  & Highway Network \cite{srivastava2015training} \\ 
    93.45 & ELU Network \cite{clevert2015fast} (18 layers) \\ \hline
    93.57 & ResNet (110 layers) \cite{he2015deep} \\ 
%96.53 & Fractional Max-pooling \cite{graham2014fractional}
94.16 & LSUV Initialization Method \cite{DBLP:journals/corr/MishkinM15} \\  \hline
94.77 & Stochastic Depth \cite{huang2016deep} \\
95.24 & Swapout \cite{singhHF2016swapout} (32 layers; 64 fm) \\ \hline
95.38 & Pre-act. ResNet \cite{he2016identity} (1001 layers) \\ 
95.59 & All Convolutional Network \cite{springenberg2014striving}  \\ \hline
95.50 & Frac. Max-pool \cite{graham2014fractional} (1 test) \\
96.53 & Frac. Max-pool  \cite{graham2014fractional} (100 tests) \\ \hline
95.83 & Wide ResNet
\cite{zagoruyko2016wide} (28 layers; 160 fm) \\ \hline \hline
     95.60 & Our CRMN (28 layers; 160 fm), RRLR  \\
     \textbf{95.84} & Our CRMN (32 layers; 192 fm), RRLR  \\ \hline
    \end{tabular}
\caption{CIFAR-10 Accuracies.}
\label{tab:cifar10}
 \end{minipage}%
%\end{table}
%
\end{table}

The Street View House Numbers\cite{netzer2011reading} (SVHN) dataset is a standard benchmark involving the classification of real-world street numbers.  We ran the best performing architectures starting with fewer than 64 feature maps on CIFAR-10 and CIFAR-100 on the SVHN dataset, and report the result of testing with the model that had the best valid set performance, which had 56 layers; 48 feature maps. Note that other than using a validation set to determine an optimum learning schedule, we used the best performing models as is, with no other hyperparameter optimization or architecture changes whatsoever.
The only data augmentation used was padding the image with a margin of 4 on each side, followed by taking a random \(32\times32\) crop (as done for both CIFAR-10 and CIFAR-100). The cropped \(32\times32\) images were converted to the range \([0, 1]\), but unlike in our CIFAR experiments, we used Global Contrast Normalization, as in \cite{goodfellow2013multi}. This CRMN yielded 98.32\% test accuracy.
The recently proposed stochastic depth approach in \cite{huang2016deep} reports %100-1.75=
98.25\%.
The top result that we are currently aware of for this benchmark is 98.36\% \cite{zagoruyko2016wide}. Their model not using dropout, having 16 layers and using 64 feature maps yielded 98.15\% accuracy. Dropout was able to boost their best model by .21\%.
% A similar model without the LSTM yielded 98.20%.
%\subsection{Pre-processing}

\section{Discussion and Conclusions}

We have explored a novel deep convolutional network architecture that is capable of selectively identifying features to remember throughout the layers of a CNN. Our formulation exploits the well known property of LSTMs to both allow gradient information to be propagated backwards for many steps and remember features derived from inputs over many processing steps. This formulation also allows CNNs to be extended with a parallel network taking intermediate representations as input and subjecting them to alternative algorithmic manipulations. 

Recent work \cite{DBLP:journals/corr/CooijmansBLC16} has successfully extended Batch Normalization to LSTMs. The applicability of this method to the proposed model needs to be explored and is likely to further boost performance.
Recently proposed exponential linear units or ELUs \cite{clevert2015fast} have demonstrated superior performance to widely used rectified linear units. Here again the use of ELUs in the convolutional layers of our model have great promise to further enhance performance.
Recently proposed LSUV initialization \cite{DBLP:journals/corr/MishkinM15} has shown an improvement in performance simply by changing the way in which weights of a CNN have been initialized. The applicability of this method in the context of the proposed model needs to be examined.

Our approach has yielded particularly promising results on CIFAR-100% in terms of both performance and computational load
. As this standard benchmark correlates well with the near definitive ImageNet 2012 visual recognition challenge, as future work we intend to run an ImageNet 2012 evaluation using the insights obtained from the experiments here and the recent advances above as a guide.

\subsubsection*{Acknowledgments}
We thank Samsung for supporting this research.

%\section*{References}

\small  
\bibliography{resmemnet}
  
\medskip

\small

\end{document}